\documentclass[10pt,twocolumn,letterpaper]{article}

\usepackage{multirow}
\usepackage{colortbl}
\usepackage{cuted}
\usepackage{pifont} 
\usepackage{algorithmic}
\usepackage{algorithm}
\usepackage[pagenumbers]{iccv} 

\definecolor{iccvblue}{rgb}{0.21,0.49,0.74}
\usepackage[pagebackref,breaklinks,colorlinks,allcolors=iccvblue]{hyperref}

\def\paperID{03395} 
\def\confName{ICCV}
\def\confYear{2025}

\title{StruMamba3D: Exploring Structural Mamba for Self-supervised \\ Point Cloud Representation Learning}

\author{Chuxin Wang$^{1,2}$, Yixin Zha$^{1}$, Wenfei Yang$^{1,2,}$\thanks{Corresponding author}, Tianzhu Zhang$^{1,2}$\\
$^1$University of Science and Technology of China \\
$^2$National Key Laboratoray of Deep Space Exploration, Deep Space Exploration Laboratory\\
\{wcx0602, zyxcn\}@mail.ustc.edu.cn, \{yangwf, tzzhang\}@ustc.edu.cn
}

\begin{document}
\maketitle
\begin{strip}
	\centering
	\includegraphics[width=0.90\textwidth]{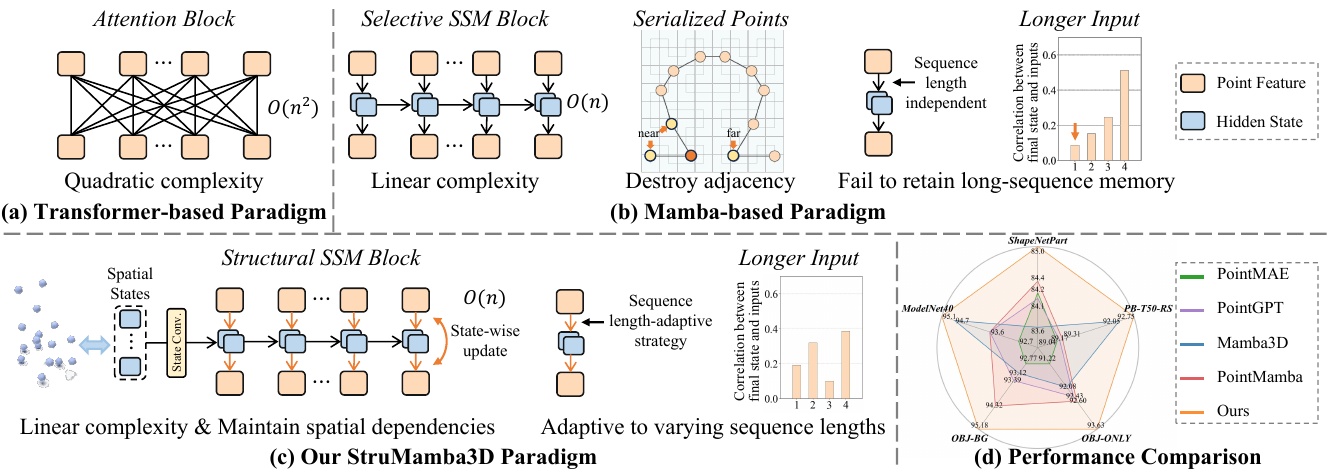}
    \vspace{-1pt}
    \captionsetup{type=figure,font=small,position=top}
    \caption{\textbf{(a):} Transformer-based paradigm uses the attention block with quadratic complexity to model the dependencies between points.
    \textbf{(b):} Mamba-based paradigm uses the selective SSM with linear complexity. However, serialized points destroy the adjacency of 3D points, and the pre-trained selection mechanism fails to retain long-sequence memory. 
    \textbf{(c):} Our StruMamba3D paradigm uses the structural SSM to maintain the spatial dependencies among points and the sequence length-adaptive strategy to retain long-sequence memory.
    \textbf{(d):} Our StruMamba3D significantly outperforms existing Transformer-based and Mamba-based methods.}
    \label{tissue}
    \vspace{-14pt}
\end{strip}
\begin{abstract}
    Recently, Mamba-based methods have demonstrated impressive performance in point cloud representation learning by leveraging State Space Model (SSM) with the efficient context modeling ability and linear complexity.
However, these methods still face two key issues that limit the potential of SSM: 
Destroying the adjacency of 3D points during SSM processing and failing to retain long-sequence memory as the input length increases in downstream tasks.
To address these issues, we propose StruMamba3D, a novel paradigm for self-supervised point cloud representation learning.
It enjoys several merits. 
First, we design spatial states and use them as proxies to preserve spatial dependencies among points.
Second, we enhance the SSM with a state-wise update strategy and incorporate a lightweight convolution to facilitate interactions between spatial states for efficient structure modeling.
Third, our method reduces the sensitivity of pre-trained Mamba-based models to varying input lengths by introducing a sequence length-adaptive strategy.
Experimental results across four downstream tasks showcase the superior performance of our method. 
In addition, our method attains the SOTA 95.1\% accuracy on ModelNet40 and 92.75\% accuracy on the most challenging split of ScanObjectNN without voting strategy.
\end{abstract}

\vspace{-1em}
\section{Introduction}\label{sec:itr}

Point cloud representation learning aims to extract geometric and semantic features from point clouds to support downstream tasks such as classification, segmentation, and detection.
As a foundational task in Computer Vision (CV), it is crucial for various real-world applications, including autonomous driving~\cite{li2020deep, chen20203d}, AR~\cite{guo2020deep}, and robotics~\cite{pomerleau2015review, wang2023long}. 
However, unlike images or text with structured arrangements, point clouds are inherently unordered and sparse, presenting challenges for designing effective models.

To solve this problem, numerous point cloud representation learning methods have been proposed, 
which can be broadly divided into three categories according to model architectures: Point-based methods~\cite{qi2017pointnet, qi2017pointnet++, li2018pointcnn, wang2019dgcnn, thomas2019kpconv, qian2022pointnext}, Transformer-based methods~\cite{guo2021pct, yu2022pointbert, liu2022maskpoint, pang2022pointmae, zhang2022pointm2ae, zha2025exploring}, and Mamba-based methods~\cite{liang2024pointmamba, han2024mamba3d, zhang2024pcm}.
Point-based methods directly process point clouds by modeling the relationships between individual points, such as through set abstraction~\cite{qi2017pointnet, qi2017pointnet++} and graph convolution~\cite{li2018pointcnn, wang2019dgcnn}. 
These methods offer a simple and computationally efficient design. 
However, their limited receptive field hinders capturing global information, constraining complex structure modeling.
Transformer-based methods leverage attention mechanisms to capture long-range dependencies, expanding the receptive field. 
Moreover, inspired by self-supervised learning paradigms in NLP~\cite{devlin2018bert} and CV~\cite{he2022masked}, many transformer-based methods~\cite{yu2022pointbert, liu2022maskpoint, pang2022pointmae, zhang2022pointm2ae} utilize large unlabeled point cloud datasets for pre-training, achieving impressive results in downstream tasks. 
However, these methods have limited scalability due to the quadratic complexity of the attention mechanism, as shown in \cref{tissue}(a). 
To overcome this limitation, Mamba-based methods~\cite{liang2024pointmamba, han2024mamba3d, zhang2024pcm} first convert the 3D points into a 1D sequence and then utilize selective State Space Model (SSM)~\cite{gu2023mamba} with linear complexity to extract point features. 
Leveraging the robust information aggregation ability of selective SSM, Mamba-based methods have attained promising performance.
Although existing Mamba-based methods~\cite{liang2024pointmamba, han2024mamba3d, zhang2024pcm} have achieved comparable performance to transformer-based methods, two key issues limit the potential of SSM:
\textbf{(1) Distort the adjacency of points.}
As shown in \cref{tissue}(b), although methods~\cite{liang2024pointmamba, zhang2024pcm} use serialization strategies to align input points, they cannot ensure that spatially adjacent points remain neighbors in the 1D sequence.
Unlike text and images, points inherently lack context information and rely on local structures as base feature units.
This spatial distortion hampers Mamba to model fine-grained structure information.
\textbf{(2) Fail to retain long-sequence memory in downstream tasks.}
Mamba-based methods~\cite{liang2024pointmamba, han2024mamba3d} use Masked Point Modeling (MPM)~\cite{yu2022pointbert, pang2022pointmae} as a pre-training task, which creates a significant discrepancy in input sequence lengths between pre-training and downstream tasks.
Mamba leverages a selection mechanism to inductively learn new information while selectively forgetting outdated knowledge.
However, the selection mechanism does not account for variations in input sequence length.
Consequently, when Mamba is pre-trained on short sequences, it tends to update states more frequently.
As the input sequence length increases in downstream tasks, frequent state updates lead to Mamba struggling to retain long-sequence memory, as shown in \cref{tissue}(b).
The deficiency in long-sequence memory adversely affects the long-range semantics modeling, resulting in suboptimal performance.
To unleash the potential of Mamba for point cloud representation learning, we explore solutions to address the above two issues.
\textbf{(1) How to maintain spatial dependencies among points?} 
Maintaining spatial relationships among 3D points within a 1D input sequence is challenging.
Mamba efficiently aggregates contextual features from the 1D input sequence via hidden states.
Is it possible to design a structural Mamba that models spatial relationships among points through hidden states?
\textbf{(2) How to retain long-sequence memory for longer inputs?} 
When handling longer input sequences, states need to retain memory for an extended period to effectively capture the contextual information across the entire sequence.
If the model adjusts the frequency of state updates when the sequence length changes, it can preserve memory for longer sequences.

To achieve the above goals, we propose a novel StruMamba3D paradigm for self-supervised point cloud representation learning, which consists of two core components: structural SSM block and sequence length-adaptive strategy.
\textbf{In structural SSM block}, we leverage states in SSM to model the spatial relationships among points.
We begin by assigning positional attributes to the states, using them as proxies to capture the local structure of point clouds.
Next, we propose a state-wise update strategy that incorporates the positional relationship between input points and spatial states during the state update and propagation process.
Based on spatial states and state-wise update strategy, we can maintain spatial dependencies among points during the SSM process.
To address the limitation of standard Mamba ignoring inter-state interactions, we introduce a lightweight convolution module for spatial states.
\textbf{In sequence length-adaptive strategy}, 
the adaptive state update mechanism is first employed to adjust the state update frequency based on sequence length, ensuring that the states retain long-sequence memory as the sequence length grows.
Additionally, we introduce a spatial state consistency loss in the pre-training task, which enforces consistency in the updated states across inputs of varying sequence lengths.
The spatial state consistency loss enhances the robustness of the Mamba-based pre-trained model to varying input lengths.

To sum up, the main contributions of this work are:
(1) We propose the novel StruMamba3D for self-supervised point cloud representation learning, which is the first to model the structure information by latent states in SSM.
(2) We design the structural SSM to maintain spatial dependencies among points during state information selection and propagation, while the sequence length-adaptive strategy reduces the sensitivity of pre-trained Mamba-based models to varying input lengths.
(3) Experimental results on four downstream tasks showcase the superior performance of our method. Additionally, comprehensive ablation studies validate the effectiveness of our designs.

\section{Related Work}\label{sec:rw}

We briefly review methods for point cloud representation learning and self-supervised representation learning.

\noindent\textbf{Point Cloud Representation Learning.}
To effectively leverage the sparse, unordered and irregular point cloud, various point cloud representation learning methods have been proposed, which can be broadly categorized into three groups:
Point-based methods~\cite{qi2017pointnet, qi2017pointnet++, li2018pointcnn, wang2019dgcnn, thomas2019kpconv, qian2022pointnext}, Transformer-based methods~\cite{guo2021pct, yu2022pointbert, liu2022maskpoint, pang2022pointmae, zhang2022pointm2ae, wang2024rethinking}, and Mamba-based methods~\cite{liang2024pointmamba, han2024mamba3d, zhang2024pcm}.
Point-based methods directly process raw point clouds to learn geometric features.
For example, PointNet~\cite{qi2017pointnet} employs MLPs to each point independently, followed by global max pooling to aggregate features.
PointNet++~\cite{qi2017pointnet++} extends PointNet by introducing a hierarchical architecture for multi-scale feature extraction.
Subsequent methods~\cite{li2018pointcnn, wang2019dgcnn, thomas2019kpconv, qian2022pointnext,ma2022pointmlp} explore convolutional techniques for point clouds.
%
%
DGCNN~\cite{wang2019dgcnn} leverages graph convolution to model structural information among neighboring points, while KPConv~\cite{thomas2019kpconv} introduces a convolution method based on dynamic kernel points to improve feature learning.
To capture long-range dependencies, methods~\cite{guo2021pct, zhao2021pt1, wu2022pt2, wu2024pt3} have explored transformer architectures for point clouds. 
%
%
PCT~\cite{guo2021pct} applies global attention to points, while Point Transformer~\cite{zhao2021pt1} introduces local attention to capture local geometry features. 
Recent variants~\cite{wu2022pt2, wu2024pt3} further refine the architectures to improve performance and efficiency. 
These supervised methods~\cite{thomas2019kpconv, qian2022pointnext, guo2021pct, zhao2021pt1, zhang2024pcm} have demonstrated outstanding performance in specific tasks. 
%
However, these methods are inherently limited by the specific data domains on which they are trained, leading to poor generalization to other domains.
%
%

\noindent\textbf{Self-supervised Representation Learning.}
To solve the above issues and fully utilize a large number of unlabeled data, self-supervised representation learning methods~\cite{xie2020pointcontrast, zhang2021crosscontrast, yu2022pointbert, pang2022pointmae, liu2022maskpoint, zhang2022pointm2ae, chen2023pointgpt} have emerged. 
Some contrastive learning-based methods~\cite{xie2020pointcontrast, zhang2021crosscontrast} train models to learn distinguishing features by differentiating between positive and negative samples. 
For instance, PointContrast~\cite{xie2020pointcontrast} leverages feature consistency across different views of the same point cloud.
Inspired by the success of BERT~\cite{devlin2018bert} and MAE~\cite{he2022masked}, many transformer-based methods~\cite{yu2022pointbert, pang2022pointmae, liu2022maskpoint, zhang2022pointm2ae, chen2023pointgpt} adopt MPM task for pre-training, encouraging models to infer masked patches from visible ones.
PointBERT~\cite{yu2022pointbert} predicts discrete tokens for masked points, while PointMAE~\cite{pang2022pointmae} reconstructs masked point clouds. 
Additionally, recent studies~\cite{zhang2023i2p, qi2023recon} integrate cross-modal information, such as images and text, to enhance pre-training.
However, transformer-based methods suffer from quadratic complexity, posing challenges for handling long sequences and making them unsuitable for resource-constrained devices.
Recently, SSMs~\cite{kalman1960new, gu2021efficiently, gu2021combining, gu2023mamba, dao2024mamba2, wang2025state} have emerged as a prominent research focus due to their linear complexity and powerful information aggregation ability.
%
In the field of point cloud representation learning, several Mamba-based methods~\cite{liang2024pointmamba, han2024mamba3d, zhang2024pcm} have been proposed. 
PointMamba~\cite{liang2024pointmamba} employs space-filling curves to serialize point clouds, establishing a simple yet effective baseline.
PCM~\cite{zhang2024pcm} introduces multiple serialization strategies and a bidirectional scanning mechanism to capture comprehensive point cloud structures, but also introduces larger computational overhead.
Mamba3D~\cite{han2024mamba3d} also leverages a bidirectional scanning mechanism and proposes a local norm pooling operation to enhance local geometric features.
However, these methods introduce structural distortions when converting 3D points into 1D sequences, limiting model performance, especially in tasks requiring fine-grained geometry like part segmentation.
Moreover, they neglect the impact of varying input lengths between pre-training and downstream tasks.
Unlike these methods, we propose StruMamba3D, which leverages states as proxies to maintain spatial dependencies among points.
Additionally, we introduce a sequence-length adaptive strategy to ensure that the states retain long-sequence memory as the sequence length grows in downstream tasks.
%

\section{Method}\label{sec:md}

Below, we review the previous SSMs (\cref{preliminaries}), followed by an overview of our StruMamba3D (\cref{overview}).
Then, we introduce the details of Structural SSM Block (\cref{ssm}) and Sequence Length-adaptive Strategy (\cref{pretraining}).

\begin{figure*}[!t]
    \vspace{-0.5em}
    \begin{center}
        \includegraphics[width=0.93\textwidth]{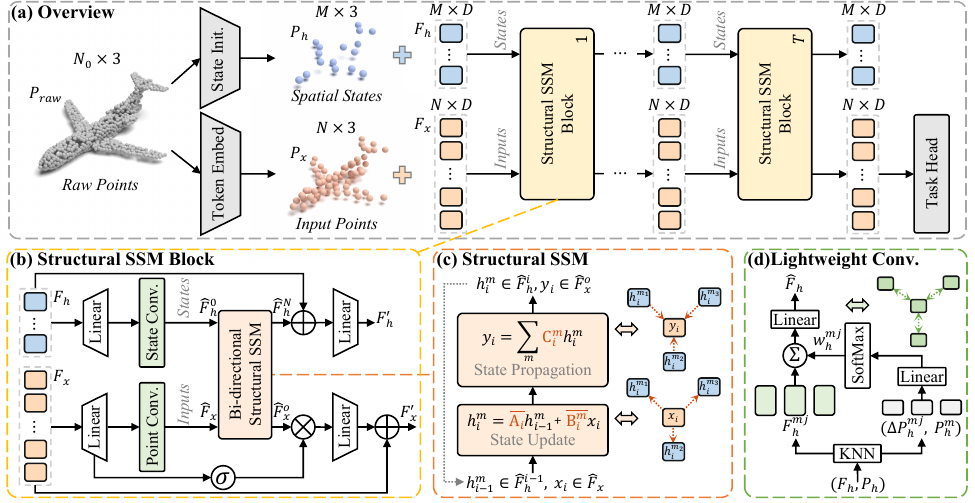}
        \vspace{-1em}
        \caption{\textbf{Overview of StruMamba3D.} 
        \textbf{(a):} Architecture of StruMamba3D.
        \textbf{(b):} Details of Structural SSM Block. 
        We employ structural SSM to capture dependencies between spatial states and input points.
        Additionally, we integrate a lightweight convolution module to introduce local interactions among spatial states or input points.
        \textbf{(c):} Details of Structural SSM.
        \textbf{(d):} Details of Lightweight Convolution.
        }
        \label{fig:overview}
        \vspace{-2.0em}
    \end{center}
\end{figure*}

\subsection{Preliminaries}
\label{preliminaries}
State Space Model (SSM) is a continuous system that maps the input $x(t)$ to $y(t)$ through implicit states $h(t) \in \mathbb{R}^M$:
\begin{equation}
    \label{ssm_1}
    h'(t) = \mathbf{A} h(t) + \mathbf{B} x(t), \ y(t) = \mathbf{C} h'(t),
\end{equation}
where the matrix $\mathbf{A}$ governs the evolution of states, while the matrices $\mathbf{B}$ and $\mathbf{C}$ are used for modeling the relationship between inputs and outputs through the latent states.
To handle discrete-time sequence data, the zero-order hold discretization rule is used:
\begin{equation}
    \label{ssm_2}
    \begin{gathered}
        h_t = \mathbf{\overline{A}} h_{t-1} + \mathbf{\overline{B}} x_t, \ y_t = \mathbf{C} h_t,\\
        \mathbf{\overline{A}} = \exp (\mathbf{\Delta} \mathbf{A}), \ \mathbf{\overline{B}} =(\mathbf{\Delta} \mathbf{A})^{-1} (\exp (\mathbf{\Delta} \mathbf{A}) - \mathbf{I}) \mathbf{\Delta} \mathbf{B},
    \end{gathered}
\end{equation}
where $\mathbf{\Delta}$ represents the sampling interval between consecutive inputs $x_{t-1}$ and $x_t$.
%
%
However, due to the linear time-invariant nature of SSM, the parameters $(\mathbf{\Delta}, \mathbf{A}, \mathbf{B}, \mathbf{C})$ remain fixed across all time steps, which limits the expressive capacity of SSM.
To overcome this limitation, Mamba~\cite{gu2023mamba} treats the parameters $(\mathbf{\Delta}, \mathbf{B}, \mathbf{C})$ as functions of input $x_t$, transforming the SSM into a time-varying model:
\begin{equation}
    \label{ssm_4}
    h_t = \phi_\mathbf{\overline{A}}(x_t) h_{t-1} + \phi_\mathbf{\overline{B}}(x_t) x_t, \ y_t = \phi_\mathbf{C}(x_t) h_t,
\end{equation}
where $\phi_*(x_t)$ denotes the linear projection of input $x_t$.
Additionally, Mamba introduces a hardware-aware scan algorithm to achieve near-linear complexity.
Our StruMamba3D adopts the same hardware-aware scan algorithm, maintaining linear complexity while exhibiting strong structural modeling capabilities for point clouds.
%
%

\subsection{Overview}
\label{overview}
Despite Mamba achieving remarkable results on sequential data, maintaining the spatial relationships among 3D points remains challenging.
We aim to model the spatial relationships by latent states and design StruMamba3D for point cloud representation learning, as illustrated in~\cref{fig:overview}(a).
Given raw point cloud $P_{raw} \in \mathbb{R}^{N_0 \times 3}$, we obtain the input points $P_{x} \in \mathbb{R}^{N \times 3}$ using farthest point sampling (FPS) and $K$-nearest neighbor clustering (KNN).
Then, we use a lightweight PointNet~\cite{qi2017pointnet} to extract the input token embeddings $F_{x} \in \mathbb{R}^{N \times D}$.
Meanwhile, we initialize the spatial states $P_{h} \in \mathbb{R}^{M \times 3}$ as the proxies for local structures from the raw point cloud.
Finally, the spatial states and input token embeddings are fed into a feature encoder composed of multiple structural SSM blocks, where the spatial states act as a bridge to connect the inputs and outputs.
Notably, we use the output spatial states from the previous structural SSM block as the input spatial states for the current block.

%
%

\subsection{Structural SSM Block}
\label{ssm} 
The structural SSM block is designed to capture complex structural features by preserving spatial dependencies among points throughout SSM processing.
As shown in~\cref{fig:overview}(b), the block takes spatial states $F_h$ and token embeddings $F_x$ as input and produces updated spatial states $F'_h$ and token embeddings $F'_x$ as output.
To clarify the details of structural SSM block, we describe each component in sequence: Spatial State Initialization, State-wise Update Mechanism, Structural SSM, and Lightweight Convolution.


\noindent\textbf{Spatial State Initialization.}
In original Mamba, latent states do not contain geometric information, making it difficult to model the local structure of point clouds. 
To address this, we introduce positional attributes $P_{h} \in \mathbb{R}^{M \times 3}$ to the latent states, enabling each state to focus on distinct regions of the point cloud.
Due to the sparsity of point clouds, we initialize state positions $P_{h}$ based on raw points $P_{raw}$ to ensure comprehensive coverage of point clouds. 
Specifically, we employ FPS and KNN clustering to segment the point cloud into groups $\{\mathcal{G}_m\}_{m=1}^M$, calculating the centroid:
\begin{equation}
    \label{eq:centroid}
    P_{h}^m = \frac{1}{|\mathcal{G}_m|} \sum_{P_i \in \mathcal{G}_m} P_i,
\end{equation}
where $P_{h}^m$ represents the positional attribute of the $m$-th state.
We refer to states with positional attributes as spatial states and then use linear mapping $\phi_h$ to embed positional attributes as the initial state features.
\begin{equation}
    \label{eq:state_init}
    F_{h} = \phi_h(P_{h}) \in \mathbb{R}^{M \times D}.
\end{equation}
%
%
The state features $F_{h}$ are then used as the initial latent states in SSM and updated along with input token embeddings.

\noindent\textbf{State-wise Update.} 
As illustrated in Eq.~\eqref{ssm_4}, the state update and propagation equations of the original Mamba rely solely on input tokens, disregarding the spatial relationships between input points and spatial states.
This limitation makes it challenging for Mamba to effectively model the structural information of point clouds.
To address this, we modify the state update and propagation equations to explicitly model the spatial dependencies between states and inputs.
Specifically, we first calculate the relative offsets between the input points $P_x$ and the state positions $P_h$:
\begin{equation}
    \label{eq:offset}
    \triangle P_i^m = P_{x}^i - P_{h}^m,
\end{equation}
where $\triangle P_i^m$ denotes the relative offset between the $i$-th input point and the $m$-th state.
Next, we map these offsets to the SSM parameters $(\mathbf{B}, \mathbf{C})$, which govern state update and propagation.
It is important to note that considering only spatial relationships is insufficient, as the significance of information varies across different points.
Thus, we design the following state update and propagation mechanism:
%
%
\begin{equation}
    \label{eq:ssm_param_mod}
    (\mathbf{B}_i^m, \mathbf{C}_i^m) = \operatorname{\phi}(x_i) + \operatorname{MLP}(\triangle P_i^m),
\end{equation}
where $\operatorname{MLP}$ represents a multi-layer perceptron, $\operatorname{\phi}$ is a linear projection and $x_i$ denotes the feature of the $i$-th input point.
Based on the modified parameters, the spatial states can selectively update features using points within the same regions, while the input points can acquire local structural information from the states, as shown in~\cref{fig:overview}(c).

\noindent\textbf{Structural SSM.} 
By leveraging spatial states and state-wise updates, we effectively model the structural information of point clouds.
However, due to unidirectional scanning mechanism, bidirectional information exchange between input points cannot be achieved within a single forward pass.
To address this, we adopt a bidirectional scanning mechanism inspired by VisionMamba~\cite{pang2022pointmae}.
Specifically, we apply both forward and backward structural SSMs, with the backward SSM processing input points in reverse order.
Both use spatial states $\hat{F}_{h}$ as initial states $\hat{F}_{h}^0$, generating updated features for input points and spatial states.
Finally, we fuse these outputs using a linear layer $\phi_o$ to obtain the final features $F'_{x}$ and $F'_{h}$ for input points and spatial states.
\begin{equation}
    \label{eq:ssm_fusion}
    F'_{x}, F'_{h} = \phi_o (\operatorname{SSM}_{f}(\hat{F}_{x}, \hat{F}_{h}) + \operatorname{SSM}_{b}({\hat{F}_{x}},\hat{F}_{h})),
\end{equation}
where ${f}$ and ${b}$ denote the forward and backward processes.

\noindent\textbf{Lightweight Convolution.} 
The proposed structural SSM effectively models the dependencies between spatial states $P_{h}$ and input points $P_{x}$.
However, we observe that the spatial states $P_{h}$ in SSMs remain isolated, lacking direct interaction with each other.
To address this, we introduce a lightweight convolution module, as shown in~\cref{fig:overview}(d).
Inspired by graph convolution~\cite{wang2019dgcnn}, this module aggregates features from neighboring states based on relative position information.
Specifically, for each spatial state $P^m_h$, we first identify its $k$-nearest neighbors $\mathcal{N}(m)$ and compute the relative positions $\triangle P_h^{mj}, j \in \mathcal{N}(m)$.
We then generate attention weights $\{w_{mj}\}$ for each neighbor using a linear layer $\phi_w$, followed by a softmax operation:
\begin{equation}
    \label{eq:attn_weight} 
    w_h^{mj} = \text{softmax}(\phi_w (\triangle P_h^{mj}, P_h^m)).
\end{equation}
The output state feature $\hat{F}_{h}^{m}$ for each state $P^m_h$ is obtained through weighted aggregation:
\begin{equation}
    \label{eq:local_feat}
    \hat{F}_{h}^{m} = \phi_c (\sum_{j \in \mathcal{N}(m)} w_h^{mj}F_h^{mj}),
\end{equation}
where $F_h^{mj}$ denotes the neighboring state features and $\phi_c$ is a linear layer.
By extending the receptive field of spatial states, this lightweight convolution enhances their ability to capture global semantic information. 
Furthermore, we replace the causal 1D convolution in the original Mamba block with this module, making it better suited for processing input points with complex geometric structures.

\subsection{Sequence Length-adaptive Strategy}
\label{pretraining}
To address the issue of insufficient long-sequence memory in downstream tasks, we propose a sequence length-adaptive strategy that integrates an adaptive state update mechanism and spatial state consistency loss.

\noindent\textbf{Adaptive State Update Mechanism.}
\begin{figure}[!t]
    \vspace{-0.5em}
    \begin{center}
        \includegraphics[width=0.47\textwidth]{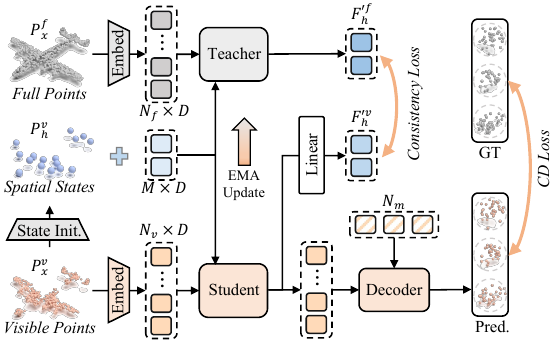}
        \vspace{-1em}
        \caption{
            \textbf{Overview of the pre-training framework.} 
            We pre-train the StruMamba3D (student model) using the MPM task and the spatial state consistency loss.
            The teacher model has the same structure as the student model, but its parameters are updated using the exponential moving average strategy.
            } 
        \label{fig:pretraining}
    \end{center}
    \vspace{-2em}
\end{figure}
According to Eq.~\eqref{ssm_2}, the parameter sampling interval $\mathbf{\Delta}$ controls the frequency of state updates by influencing the parameters $(\mathbf{\overline{A}}, \mathbf{\overline{B}})$ in SSMs.
A larger $\mathbf{\Delta}$ causes the model to update state features more frequently, whereas a smaller $\mathbf{\Delta}$ allows the model to maintain its current state with minimal changes. 
To preserve memory for longer sequences in downstream tasks, we propose an adaptive state update mechanism. 
This mechanism adjusts the sampling interval $\mathbf{\Delta}$ according to the input sequence length, allowing the model to maintain a consistent total sampling time $\mathbf{\Delta}_{all} = \sum_{i=1}^{N} \mathbf{\Delta}_i$ across different sequence lengths.
Specifically, we design a learnable parameter $\tau$, which regulates the total sampling time $\mathbf{\Delta}_{all}$ for sequences of varying lengths. 
The sampling interval $\mathbf{\Delta}_i$ for each token in the inputs is calculated as follows:
\begin{equation}
    \label{eq:adaptive_state_update}
    \mathbf{\Delta}_i = \frac{\tau \times \mathbf{\Delta}_{i}}{\sum_{i=1}^{N} \mathbf{\Delta}_{i}}.
\end{equation}
Based on the adaptive state update mechanism, the model can maintain a consistent total sampling time $\mathbf{\Delta}_{all} = \tau$ across different sequence lengths.

\begin{table*}[!t]
    \centering
    \vspace{-1em}
    \caption{\textbf{Shape Classification on ScanObjectNN and ModelNet40 Datasets.} For ScanObjectNN, we report the classification accuracy(\%) over the three subsets: OBJ-BG, OBJ-ONLY, and PB-T50-RS. And $\dagger$ indicates using simple rotational augmentation~\cite{dong2023act} for fine-tuning.}
    \vspace{-0.9em}
    \small
    \label{table:cls}
    \setlength{\tabcolsep}{4.4pt}
    \begin{tabular*}{0.98\textwidth}{l|ccc|ccc|cc}
        \toprule
        \multirow{2}*{Method} & \multirow{2}*{Backbone} & \multirow{2}*{Param. (M)} & \multirow{2}*{FLOPs (G)} & \multicolumn{3}{c|}{ScanObjectNN} & \multicolumn{2}{c}{MODELNET40} \\
        \cline{5-9}
        ~ &~ &~ &~ & $\text{OBJ-BG}$ & $\text{OBJ-ONLY}$ & $\text{PB-T50-RS}$ & $\text{w/o Voting}$ & $\text{w/ Voting}$ \\
        \hline
        \multicolumn{9}{c}{\textit{Supervised learning only}} \\
        \hline
        PointNet~\cite{qi2017pointnet} & - & 3.5 & 0.5 & 73.3 & 79.2 & 68.0 & 89.2  & - \\
        PointNet++~\cite{qi2017pointnet++} & - & 1.5 & 1.7 & 82.3 & 84.3 & 77.9 & 90.7  & - \\
        PointCNN~\cite{li2018pointcnn} & - & 0.6 & - & 86.1 & 85.5 & 78.5 & 92.2  & - \\
        DGCNN~\cite{wang2019dgcnn} & - & 1.8 & 2.4 & 82.8 & 86.2 & 78.1 & 92.9  & - \\
        PRANet~\cite{fan2020pranet} & - & 2.3 & - & - & - & 81.0 & 93.7  & - \\
        PointNeXt~\cite{qian2022pointnext} & - & 1.4 & 3.6 & - & - & 87.7 & 94.0  & - \\
        PointMLP~\cite{ma2022pointmlp} & - & 12.6 & 31.4 & - & - & 85.4 & 94.5  & - \\
        DeLA~\cite{chen2023dela} & - & 5.3 & 1.5 & - & - & 88.6 & 94.0  & - \\
        PCM~\cite{zhang2024pcm} & - & 34.2 & 45.0 & - & - & 88.1 & 93.4  & - \\
        \hline
        \multicolumn{9}{c}{\textit{Pre-training using single-modal information}} \\
        \hline
        PointBERT~\cite{yu2022pointbert} & Transformer & 22.1 & 4.8 & 87.43 & 88.12 & 83.07 & 92.7 & 93.2 \\
        MaskPoint~\cite{liu2022maskpoint}& Transformer & 22.1 & 4.8 & 89.30 & 88.10 & 84.30 & - & 93.8 \\
        PointM2AE~\cite{zhang2022pointm2ae}& Transformer & 12.7 & 7.9 & 91.22 & 88.81 & 86.43 & 92.9 & 93.4 \\
        $\text{PointMAE}^\dagger$~\cite{pang2022pointmae} & Transformer & 22.1 & 4.8 & 92.77 & 91.22 & 89.04 & 92.7 & 93.8 \\
        $\text{PointGPT-S}^\dagger$~\cite{chen2023pointgpt} & Transformer & 29.2 & 5.7 & 93.39 & 92.43 & 89.17 & 93.3 & 94.0 \\
        $\text{PointMamba}^\dagger$~\cite{liang2024pointmamba} & Mamba & 12.3 & 3.1 & 94.32 & 92.60 & 89.31 & 93.6 & 94.1 \\
        $\text{Mamba3D}^\dagger$~\cite{han2024mamba3d} & Mamba & 16.9 & 3.9 & 93.12 & 92.08 & 92.05 & 94.7 & 95.1 \\
        \rowcolor{gray!20} 
        \textbf{$\text{Ours}^\dagger$} & Structural SSM & 15.8 & 4.0 & \textbf{95.18} & \textbf{93.63} & \textbf{92.75} & \textbf{95.1} & \textbf{95.4} \\
        \hline
        \multicolumn{9}{c}{\textit{Pre-training using cross-modal information}} \\
        \hline
        $\text{ACT}^\dagger$~\cite{dong2023act} & Transformer & 22.1 & 4.8 & 93.29 & 91.91 & 88.21 & 93.7 & 94.0 \\
        Joint-MAE~\cite{guo2023joint} & Transformer & 22.1 & - & 90.94 & 88.86 & 86.07 & - & 94.0\\
        $\text{I2P-MAE}^\dagger$~\cite{zhang2023i2pmae} & Transformer & 15.3 & - & 94.15 & 91.57 & 90.11 & 93.7 & 94.1\\
        $\text{ReCon}^\dagger$~\cite{qi2023recon} & Transformer & 43.6 & 5.3 & 95.18 & 93.29 & 90.63 & 94.5 & 94.7\\
        \bottomrule
    \end{tabular*}
    \vspace{-0.5em}
\end{table*}

\noindent\textbf{Spatial State Consistency Loss.}
To further ensure consistency in updated state features across different input sequence lengths, we introduce a spatial state consistency loss to the pre-training task.
As shown in~\cref{fig:pretraining}, we present an overview of the pre-training framework. 
Following the previous works~\cite{pang2022pointmae,liang2024pointmamba,han2024mamba3d}, we reconstruct the points in the masked regions and use the reconstruction loss $\mathcal{L}_{cd}$ to guide the model to learn discriminative features from the visible points.
For the spatial state consistency loss, we first use a teacher model to update the spatial states with the full input tokens $P_{x}^f \in \mathbb{R}^{N_{f} \times 3}$.
The state features $F'^f_{h}$ output by the teacher model serve as pseudo-labels.
We then enforce a consistency loss $\mathcal{L}_{ssc}$ between the updated state features of the student model $F'^v_{h}$ and the pseudo-labels $F'^f_{h}$:
\begin{equation}
    \label{eq:ssc_loss}
    \mathcal{L}_{ssc} = \operatorname{Smooth\ L1}(F'^v_{h}, F'^f_{h}).
\end{equation}
The spatial state consistency loss $\mathcal{L}_{ssc}$ not only enhances the robustness of our StruMamba3D to varying input sequence lengths but also improves the ability to infer the structural information of the complete point cloud from incomplete point clouds.
The final pre-training loss is:
\begin{equation}
    \label{eq:total_loss}
    \mathcal{L}_{total} = \mathcal{L}_{cd} + \lambda \times \mathcal{L}_{ssc},
\end{equation}
where $\lambda$ is the weight of the spatial state consistency loss.

\section{Experimental Results}\label{sec:exp}

\subsection{Implementation Details}
%
%
%
%
Following the previous works~\cite{pang2022pointmae, liang2024pointmamba, han2024mamba3d}, we employ $T$=12 structural SSM blocks, each with a feature dimension of $D$=384.
The number of spatial states $M$ is set to 16.
For the lightweight convolution in the structural SSM block, we set the number of neighbors to 4 for spatial states and 8 for input points.
Following previous works~\cite{yu2022pointbert, pang2022pointmae, zhang2022pointm2ae}, we pre-train the proposed model on ShapeNet dataset~\cite{chang2015shapenet}, which contains 52472 unique 3D models across 55 common object categories.
Each input raw point cloud, containing $N_0$=1024 points, is divided into 64 patches with each consisting of 32 points.
A random masking ratio of 0.6 is applied, with the decoder composed of 4 standard Mamba blocks, and the $\lambda$ in the pre-training loss is set to 2.
%
%
More details can be found in the supplementary material.
%
%
%

\subsection{Comparison on Downstream Tasks}
We present the experimental results of our StruMamba3D across various downstream tasks and compare them with previous state-of-the-art (SOTA) methods.

\noindent\textbf{Shape classification.}
ScanObjectNN \cite{yi2016scalable} is a real-world dataset containing roughly 15000 objects across 15 categories, some of which have cluttered backgrounds or partial scans.
For this dataset, we apply rotation as a data augmentation strategy~\cite{dong2023act} and use 2048 points as input.
ModelNet40 \cite{wu2015modelnet40} is a synthetic dataset comprising of 12311 clean 3D CAD models from 40 categories.
For this dataset, we use 1024 points as input and apply scale and translation augmentations~\cite{qi2017pointnet}.
As shown in~\cref{table:cls}, we conduct classification experiments on both datasets.
Compared to transformer-based methods, our model achieves significant performance improvements across multiple metrics on both datasets.
Even when compared to the best PointGPT-S~\cite{chen2023pointgpt}, our model outperforms it by 3.58\% on ScanObjectNN and 1.8\% on ModelNet40.
Additionally, on the ScanObjectNN dataset, StruMamba3D surpasses PointMamba~\cite{liang2024pointmamba} by 0.86\%, 1.03\%, and 3.44\% across the three splits, while also outperforming Mamba3D~\cite{han2024mamba3d} by 2.06\%, 1.55\%, and 0.70\%.
Furthermore, on the ModelNet40 dataset, StruMamba3D exceeds PointMamba by 1.5\% and outperforms Mamba3D by 0.4\%.
Overall, these results demonstrate that StruMamba3D outperforms all existing unimodal architectures, achieving multiple SOTA results and highlighting the effectiveness of structural modeling within Mamba.

\noindent\textbf{Part Segmentation on ShapeNetPart.}
\begin{table}[!t]
    \vspace{-1.2em}
    \begin{center}
        \setlength\tabcolsep{8pt}
        \caption{\textbf{Part Segmentation on ShapeNetPart.} We report the mean IoU for all classes (mIoU$_c$) and all instances (mIoU$_i$).}
        \small
        \label{table:ShapeNetPart}
        \vspace{-0.9em}
        \begin{tabular}{lccc}
            \toprule
            {Method}           &   Architecture                      & mIoU$_c$           & mIoU$_i$                          \\
            \hline
            \multicolumn{4}{c}{\textit{Supervised learning only}} \\
            \hline
            PointNet~\cite{qi2017pointnet} & Single-scale & 80.4 & 83.7 \\
            PointNet++~\cite{qi2017pointnet++} & Multi-scale & 81.9 & 85.1 \\
            APES~\cite{wu2023attention} & Multi-scale & 83.7 & 85.8 \\
            DeLA~\cite{chen2023dela} & Multi-scale & 85.8 & 87.0 \\
            PCM~\cite{zhang2024pcm} & Multi-scale & 85.3 & 87.0 \\
            \hline
            \multicolumn{4}{c}{\textit{Pre-training using single-modal information}} \\
            \hline
            MaskPoint~\cite{liu2022maskpoint} & Single-scale & {84.6} & 86.0 \\
            PointBERT~\cite{yu2022pointbert} & Single-scale & 84.1 & 85.6 \\
            PointMAE~\cite{pang2022pointmae} & Single-scale & 84.2 & 86.1 \\
            PointM2AE~\cite{zhang2022pointm2ae} & Multi-scale & 84.9 & 86.5 \\
            PointGPT-S~\cite{chen2023pointgpt} & Single-scale & 84.1 & 86.2 \\
            PointMamba~\cite{liang2024pointmamba} & Single-scale & 84.4 & 86.2 \\
            Mamba3D~\cite{han2024mamba3d} & Single-scale & 83.6 & 85.6 \\
            \rowcolor{gray!20} 
            \textbf{Ours} & Single-scale & \textbf{85.0} & \textbf{86.7} \\
            \bottomrule
        \end{tabular}
    \end{center}
    \vspace{-2.8em}
\end{table}
ShapeNetPart~\cite{yi2016scalable} dataset comprises 16880 models spanning 16 shape categories and annotated with 50 part labels.
Compared to classification tasks, part segmentation poses a greater challenge, as it requires assigning a label to each point, making it highly dependent on the intrinsic structural information of point clouds.
\cref{table:ShapeNetPart} presents our part segmentation results for all classes and instances on the ShapeNetPart dataset.
For single-scale models, our method achieves substantial performance improvements, surpassing Mamba3D~\cite{han2024mamba3d} by 1.4\% in mIoU$_c$ and 1.2\% in mIoU$_i$, and outperforming PointMamba~\cite{liang2024pointmamba} by 0.6\% in mIoU$_c$ and 0.5\% in mIoU$_i$.
Notably, Mamba3D without a serialization strategy performs poorly, falling behind the baseline PointMAE~\cite{pang2022pointmae}, whereas PointMamba, which incorporates serialization, exhibits improved performance.
This observation suggests the crucial role of serialization in enhancing structural modeling within Mamba-based architectures.
In contrast, our StruMamba3D leverages spatial states to effectively capture structural information, thereby eliminating the need for serialization and achieving 85.0\% in mIoU$_c$ and 86.7\% in mIoU$_i$.
Besides, StruMamba3D even outperforms the multi-scale self-supervised method PointM2AE~\cite{zhang2022pointm2ae}. 
Although multi-scale approaches typically yield better performance in part segmentation tasks, they often come with significantly higher computational costs (e.g., PCM~\cite{zhang2024pcm} requires 45.0G FLOPs).
In this work, we focus on designing an efficient and effective Mamba-based framework for structural modeling in point clouds.
While incorporating multi-scale designs could further improve part segmentation performance, such strategies may introduce heuristic dependencies.
Overall, our StruMamba3D represents a more robust and efficient Mamba-based framework for point cloud structural modeling, achieving the highest performance among all single-scale models.
%
%
%

\noindent\textbf{Few-shot Learning on ModelNet40.}
\begin{table}[!t]
    \vspace{-1.2em}
    \begin{center}
      \setlength\tabcolsep{4pt}
      \caption{\textbf{Few-Shot Classification on ModelNet40.} We report overall accuracy and standard deviation without a voting strategy.}
      \small
      \label{table:fewshot}
      \vspace{-0.9em}
      \begin{tabular}{l|cc|cc}
        \toprule
        \multirow{2}*{Method}                & \multicolumn{2}{c|}{5-way}                & \multicolumn{2}{c}{10-way}                                                                                                        \\
        \cline{2-5}
        ~                                   & $\text{10-shot}$                          & $\text{20-shot}$                          & $\text{10-shot}$                          & $\text{20-shot}$                          \\
        \hline
        PointBERT~\cite{yu2022pointbert}   & 94.6{\scriptsize $\pm$3.6}                & 93.9{\scriptsize $\pm$3.1}                & 86.4{\scriptsize $\pm$5.4}                & 91.3{\scriptsize $\pm$4.6}                \\
        MaskPoint~\cite{liu2022maskpoint}   & 95.0{\scriptsize $\pm$3.7}                & 97.2{\scriptsize $\pm$1.7}                & 91.4{\scriptsize $\pm$4.0}                & 92.7{\scriptsize $\pm$5.1}                \\
        PointMAE~\cite{pang2022pointmae}    & 96.3{\scriptsize $\pm$2.5}                & 97.8{\scriptsize $\pm$1.8}                & 92.6{\scriptsize $\pm$4.1}                & 93.4{\scriptsize $\pm$3.5}                \\
        PointM2AE~\cite{zhang2022pointm2ae} & 96.8{\scriptsize $\pm$1.8}                & 98.3{\scriptsize $\pm$1.4}                & 92.3{\scriptsize $\pm$4.5}                & 95.0{\scriptsize $\pm$3.0}                \\
        PointGPT-S~\cite{chen2023pointgpt}  & 96.8{\scriptsize $\pm$2.0}                & 98.6{\scriptsize $\pm$1.1}    & 92.6{\scriptsize $\pm$4.6}                & 95.2{\scriptsize $\pm$3.4}                \\
        PointMamba~\cite{liang2024pointmamba} & 96.9{\scriptsize $\pm$2.0} & {99.0}{\scriptsize $\pm$1.1} & 93.0{\scriptsize $\pm$4.4} & 95.6{\scriptsize $\pm$3.2} \\
        Mamba3D~\cite{han2024mamba3d} & 96.4{\scriptsize $\pm$2.2} & 98.2{\scriptsize $\pm$1.2} & 92.4{\scriptsize $\pm$4.1} & 95.2{\scriptsize $\pm$2.9} \\
        \rowcolor{gray!20} 
        \textbf{Ours} & \textbf{97.5}{\scriptsize $\pm$2.3} & \textbf{99.1}{\scriptsize $\pm$1.4} & \textbf{93.5}{\scriptsize $\pm$3.7} & \textbf{96.1}{\scriptsize $\pm$3.5} \\
        \bottomrule
      \end{tabular}
    \end{center}
    \vspace{-2.8em}
\end{table}
To evaluate the effectiveness of the proposed methods with limited fine-tuning data, we conduct few-shot classification experiments on the ModelNet40 dataset.
Following the standard procedure~\cite{yu2022pointbert, pang2022pointmae}, we perform 10 independent experiments for each setting and report the mean accuracy and standard deviation.
As shown in~\cref{table:fewshot}, our method demonstrates a significant improvement over Mamba-based methods, outperforming PointMamba~\cite{liang2024pointmamba} by 0.6\% in 5-way-10-shot and 0.5\% in 10-way-10-shot settings, and surpassing Mamba3D~\cite{han2024mamba3d} by 1.1\% and 0.9\%, respectively.
These results indicate that our method can efficiently adapt to downstream tasks with longer input sequences, even under the constraint of limited training data.
%
%

\subsection{Ablation Study}
%
%
To explore the architectural design, we perform ablation studies on the PB-T50-RS split of ScanObjectNN dataset, as well as ModelNet40 and ShapeNetPart datasets.

\noindent\textbf{Effect of Designed Modules.}
As shown in~\cref{table:ablation}, we conduct an ablation study to evaluate the effectiveness of two core modules: the Structural SSM block and the sequence length-adaptive strategy.
The baseline model is built upon the standard Mamba~\cite{gu2023mamba} block, following the same architecture as PointMAE~\cite{pang2022pointmae}.
Compared to this baseline, the Structural SSM block significantly enhances model performance, yielding improvements of 4.86\% on ScanObjectNN, 2.59\% on ModelNet40, and 2.93\% on ShapeNetPart, demonstrating its effectiveness in modeling spatial dependencies and capturing richer structural information in point clouds.
Furthermore, integrating the sequence length-adaptive strategy further enhances performance by improving the generalization to varying input lengths, thereby ensuring optimal performance across downstream tasks.
%
\begin{table}[!t]
    \begin{center}
        \small
        \setlength\tabcolsep{8pt}
        \vspace{-1.2em}
        \caption{\textbf{Effect of the designed modules.} `SLAS' indicates the sequence length-adaptive strategy used in pre-training phase.}
        \vspace{-0.9em}
        \label{table:ablation}
        \begin{tabular}{cc|ccc}
            \toprule
            Structural & \multirow{2}*{SLAS} & \multicolumn{2}{c}{Overall Accuracy} & mIoU$_c$  \\
            \cline{3-5}                 
            SSM Block & ~ & ScanNN & MN40 & SNPart\\
            \hline
            \ding{55} & \ding{55} & 87.23 & 91.86 & 81.56\\
            \ding{51} & \ding{55} & 92.09 & 94.45 & 84.49\\
            \ding{51} & \ding{51}  & \textbf{92.75} & \textbf{95.06} & \textbf{84.96}\\
            \bottomrule
        \end{tabular}
        \vspace{-2.2em}
    \end{center}
\end{table}

\noindent\textbf{Effect of Structural SSM Block.}
\begin{table}[!t]
    \begin{center}
        \small
        \setlength\tabcolsep{5pt}
        \caption{\textbf{Ablation on Structural SSM Block.} The baseline uses the standard Mamba block to replace the Structural SSM Block.}
        \vspace{-0.9em}
        \label{table:ablation_ssm_block}
        \begin{tabular}{l|ccc}
            \toprule
            \multirow{2}*{Method} & \multicolumn{2}{c}{Overall Accuracy} & mIoU$_c$  \\
            \cline{2-4}                 
            ~ & ScanNN & MN40 & SNPart \\
            \hline
            baseline & 88.24 & 92.50 & 82.08 \\
            \ \ w/ Structural SSM & 91.78 & 93.92 & 84.15 \\
            \ \ w/ Lightweight Conv. for $h$ & 92.22 & 94.65 & 84.62 \\
            \ \ w/ Lightweight Conv. for $x$ & 92.40 & 94.81 & 84.77 \\
            \ \ w/ Bidirectional Scanning & \textbf{92.75} & \textbf{95.06} & \textbf{84.96} \\
            \bottomrule
        \end{tabular}
        \vspace{-2.8em}
    \end{center}
\end{table}
As shown in~\cref{table:ablation_ssm_block}, we further validate the effectiveness of individual components within the Structural SSM block.
Starting from the baseline model with the standard Mamba block, we incrementally incorporate key modules from the Structural SSM block. 
The most significant performance gains stem from the Structural SSM and the lightweight convolution designed for spatial states, demonstrating that modeling structural information through hidden states in SSM is crucial.
Moveover, the lightweight convolution for input points and the bidirectional scanning mechanism also enhance the performance across downstream tasks.

\noindent\textbf{Effect of Structural SSM.}
Structural SSM is a crucial component of our method, as it effectively captures the structural information of point clouds through spatial states. 
To further analyze the impact of spatial state initialization and state-wise update strategy, we conduct ablation studies as shown in~\cref{table:ablation_ssm}.
Although spatial state initialization provides the states with positional information, the improvement is minimal, as Eq.~\eqref{ssm_2} does not consider spatial relationships between input points and states.
When integrating the state-wise update strategy, the model achieves a significant performance boost, with improvements of 1.42\% on ScanObjectNN, 0.94\% on ModelNet40, and 1.09\% on ShapeNetPart.
The results indicate that incorporating spatial relationships to control state information updates and propagation within the SSM is essential for effectively modeling structural information.
Moreover, we observe that relying solely on spatial relationships to generate SSM parameters results in performance degradation.
This is likely due to the varying importance of point features, which cannot be fully captured by spatial position alone.
\begin{table}[!t]
    \begin{center}
        \small
        \setlength\tabcolsep{5pt}
        \vspace{-1.2em}
        \caption{\textbf{Ablation on Structural SSM.} `SS' represents the use of spatial states in the SSM. `$\phi(x)$' denotes the SSM parameters depending on the point feature, while `$\operatorname{MLP}(\triangle P)$' represents the SSM parameters depending on the spatial relationship.}
        \vspace{-0.9em}
        \label{table:ablation_ssm}
        \begin{tabular}{ccc|ccc}
            \toprule
            \multirow{2}*{$\phi(x)$} & \multirow{2}*{SS} & \multirow{2}*{$\operatorname{MLP}(\triangle P)$} & \multicolumn{2}{c}{Overall Accuracy} & mIoU$_c$  \\  
            \cline{4-6}                 
            ~ & ~ & ~ & ScanNN & MN40 & SNPart\\
            \hline
            \ding{51} & \ding{55} & \ding{55} & 90.94 & 93.84 & 83.62\\
            \ding{51} & \ding{51} & \ding{55} & 91.33 & 94.12 & 83.87\\
            \ding{51} & \ding{51} & \ding{51} & \textbf{92.75} & \textbf{95.06} & \textbf{84.96}\\
            \ding{55} & \ding{51} & \ding{51} & 91.12 & 94.25 & 84.02\\
            \bottomrule
        \end{tabular}
        \vspace{-2.2em}
    \end{center}
\end{table}

\noindent\textbf{Effect of Sequence Length-Adaptive Strategy.}
\begin{table}[!t]
    \begin{center}
        \small
        \setlength\tabcolsep{8pt}
        \caption{\textbf{Ablation on sequence length-adaptive strategy.} `ASUM' indicates the adaptive state update mechanism, and `$\lambda$' indicates the weight of the spatial state consistency loss $\mathcal{L}_{ssc}$.}
        \vspace{-0.9em}
        \label{table:ablation_slas}
        \begin{tabular}{cc|ccc}
            \toprule
            \multirow{2}*{ASUM} & \multirow{2}*{$\mathcal{L}_{ssc}$} & \multicolumn{2}{c}{Overall Accuracy}      \\
            \cline{3-5}                 
            ~ & ~ & ScanNN & MN40 & SNPart\\
            \hline
            \ding{55} & \ding{55} & 92.09 & 94.45 & 84.49\\
            \ding{51} & \ding{55} & 92.26 & 94.57 & 84.56\\
            \ding{51} & $\lambda$ = 1 & 92.47 & 94.94 & 84.81\\
            \ding{51} & $\lambda$ = 2 & \textbf{92.75} & \textbf{95.06} & \textbf{84.96}\\
            \ding{51} & $\lambda$ = 5 & 92.57 & 94.89 & 84.77\\
            \ding{55} & $\lambda$ = 2 & 92.16 & 94.65 & 84.56\\
            \bottomrule
        \end{tabular}
        \vspace{-2.8em}
    \end{center}
\end{table} 
The sequence length-adaptive strategy consists of two key components: the adaptive state update mechanism and the spatial state consistency loss.
\cref{table:ablation_slas} presents an ablation study on these components.
When applied individually, both the adaptive state update mechanism and the spatial state consistency loss yield only marginal performance gains.
However, when used together, they lead to a significant improvement, demonstrating their complementary effects.
The adaptive state update mechanism ensures a consistent total sampling time across different input lengths, while the spatial state consistency loss ensures the consistency of spatial state features.
In addition, we perform an ablation study on the weight $\lambda$ of $\mathcal{L}_{ssc}$ and observe that the model achieves optimal performance when $\lambda$ is set to 2.


\vspace{-0.5em}
\section{Conclusion}\label{sec:clu}
\vspace{-0.5em}
We identify two key issues limiting Mamba potential: 3D point adjacency distortion during SSM processing and inadequate long-sequence memory retention in downstream tasks. 
To address these issues, we propose StruMamba3D with two core components. 
The structural SSM block maintains spatial dependencies during state selection and propagation, 
while the sequence length-adaptive strategy reduces the sensitivity of Mamba-based models to varying input lengths. 
Our method achieves substantial improvements across four downstream tasks, with ablation studies confirming the effectiveness of each module.

\vspace{-0.5em}
\section{Acknowledgment}\label{sec:ack}
\vspace{-0.5em}
This work is supported by the National Defense Basic Scientific Research program (JCKY2022911B002).

{
    \small
    \bibliographystyle{ieeenat_fullname}
    \bibliography{main}
}

\clearpage
\section*{Appendix}

\section{Overview}

In this supplementary material, we first provide additional implementation details (\cref{sec:details}). Next, we present more experimental results (\cref{sec:exp}) and qualitative analysis (\cref{sec:visual}) to validate and analyze our proposed method. Finally, we discuss the limitations of our approach and potential directions for future research (\cref{sec:discussion}).

\section{More Implementation Details}\label{sec:details}

\subsection{Structural SSM Block}
In the state-wise update strategy and sequence-length adaptive strategy, we modify the SSM parameter generation process to enhance its efficiency.
To provide a more intuitive understanding of the structural SSM block, we present its detailed operations in~\cref{algo:ssb}. 
The process begins with the normalization of input points $F_x$ and initial spatial states $F_h$ followed by their linear projection into $F_x^l$ and $F_h^l$. 
Next, a lightweight convolution is applied to $F_x^l$ and $F_h^l$, yielding $\hat{F}_x$ and $\hat{F}_h$.
Before processing $\hat{F}_x$ in both forward and backward directions, we compute the relative position offsets $\triangle P$ between the input points $P_x$ and spatial states $P_h$.
Based on the $\hat{F}_x$ and $\triangle P$, we generate the SSM parameters ($\mathbf{B}_o$, $\mathbf{C}_o$, $\mathbf{\Delta}'_o$).
A learnable parameter $\tau$ is introduced to regulate the sampling interval $\mathbf{\Delta}'_o$, producing $\mathbf{\Delta}_o$.
Utilizing $\mathbf{\Delta}_o$, we transform $\mathbf{\overline{A}}_o$, $\mathbf{\overline{B}}_o$. 
Finally, the SSM computes the output point features $\hat{F}_x^o$ and the updated spatial states $\hat{F_h^o}$. 
The forward and backward output point features, $\hat{F}_x^f$ and $\hat{F}_x^b$, are gated by $F_z$ and then summed together.
Similarly, the forward and backward spatial states, $\hat{F}_h^f$ and $\hat{F}_h^b$, are aggregated.
The final outputs, $F_x'$ and $F_h'$, are obtained by connecting the updated inputs and spatial states to $F_x$ and $F_h$ using residual connections, respectively.

\begin{algorithm*}[!t]
    \caption{Structural SSM Block Process}
    \label{algo:ssb}
    \begin{algorithmic}[1]
        \REQUIRE Input point features $F_x$: {\color{cyan}(B, N, D)}, Initial spatial states $F_h$: {\color{cyan}(B, M, D)} \\
        \ENSURE Output point features $F_x'$: {\color{cyan}(B, N, D)}, Final spatial states $F_h'$: {\color{cyan}(B, M, D)} \\
        \STATE {\color{gray}/* normalize the input point features $F_x$ */} 
        \STATE $F_x^n$: {\color{cyan}(B, N, D)} $\leftarrow$ $\operatorname{Norm}(F_x)$, $F_h^n$: {\color{cyan}(B, M, D)} $\leftarrow$ $\operatorname{Norm}(F_h)$ 
        \STATE ${F_x^l}$: {\color{cyan}(B, N, E)} $\leftarrow$ $\operatorname{Linear}^x(F_x^n)$, ${F_z^l}$: {\color{cyan}(B, N, E)} $\leftarrow$ $\operatorname{Linear}^z(F_x^n)$, ${F_h^l}$: {\color{cyan}(B, M, E)} $\leftarrow$ $\operatorname{Linear}^h(F_h^n)$
        \STATE $\hat{F}_x$: {\color{cyan}(B, N, E)} $\leftarrow$ $\operatorname{LightConv.}(F_x^l)$, $\hat{F}_h$: {\color{cyan}(B, M, E)} $\leftarrow$ $\operatorname{LightConv.}(F_h^l)$
        \STATE {\color{gray}/* process with different direction */}
        \FOR {$ o\ \text{in} \{\operatorname{forward}, \operatorname{backward}\}$}
        \STATE {\color{gray}/* $\triangle P$ is the relative offsets between the input points and spatial states: {\color{cyan}(B, N, M, 3)} */}
        \STATE $\mathbf{B}_o$: {\color{cyan}(B, N, M)} $\leftarrow$ $\operatorname{\phi_{\mathbf{B}}}(\hat{F}_x) + \operatorname{MLP_{\mathbf{B}}}(\triangle P)$
        \STATE $\mathbf{C}_o$: {\color{cyan}(B, N, M)} $\leftarrow$ $\operatorname{\phi_{\mathbf{C}}}(\hat{F}_x) + \operatorname{MLP_{\mathbf{C}}}(\triangle P)$
        \STATE {\color{gray}/* softplus ensures positive $\mathbf{\Delta}_o$ */}
        \STATE $\mathbf{\Delta'}_o$: {\color{cyan}(B, N, E)} $\leftarrow$ $\log(1 + \exp (\phi_{\mathbf{\Delta}}(\hat{F}_x)))$
        \STATE {\color{gray}/* learnable parameter $\tau$ regulates the total sampling time $\mathbf{\Delta}_{all}$ */}
        \STATE $\mathbf{\Delta}_o$: {\color{cyan}(B, N, E)} $\leftarrow$ $\tau \times \mathbf{\Delta'}_o / (\sum_{i=1}^{N} \mathbf{\Delta'}_o^{i})$
        \STATE {\color{gray}/* $\operatorname{Parameter}^{\mathbf{A}}_o$ is learnable parameter: {\color{cyan}(M, E)} */}
        \STATE $\mathbf{\overline{A}}_o$: {\color{cyan}(B, N, M, E)} $\leftarrow$ $\mathbf{\Delta}_o \otimes \operatorname{Parameter}^{\mathbf{A}}_o$
        \STATE $\mathbf{\overline{B}}_o$: {\color{cyan}(B, N, M, E)} $\leftarrow$ $\mathbf{\Delta}_o \otimes \mathbf{B}_o$
        \STATE $\hat{F}_x^o$: {\color{cyan}(B, N, E)}, $\hat{F}_h^o$: {\color{cyan}(B, M, E)} $\leftarrow$ $\mathbf{SSM}(\mathbf{\overline{A}}_o, \mathbf{\overline{B}}_o, \mathbf{C}_o)(\hat{F}_x, \hat{F}_h)$
        \ENDFOR
        \STATE {\color{gray}/* gated linear unit and residual connection */}
        \STATE $F_x'^f$: {\color{cyan}(B, N, E)} $\leftarrow$ $\hat{F}_x^f \odot \operatorname{SiLU}(F_z)$, $F_x'^b$: {\color{cyan}(B, N, E)} $\leftarrow$ $\hat{F}_x^b \odot \operatorname{SiLU}(F_z)$
        \STATE $F_x'$: {\color{cyan}(B, N, C)} $\leftarrow$ $\operatorname{Linear}^{out}(F_x'^f + F_x'^b) + F_x$
        \STATE $F_h'$: {\color{cyan}(B, M, C)} $\leftarrow$ $\operatorname{Linear}^h(\hat{F}_h^f + \hat{F}_h^b) + F_h$
        \RETURN $F_x'$ and $F_h'$
    \end{algorithmic}
\end{algorithm*}

\subsection{Pre-training Details}
Following previous works~\cite{pang2022pointmae}, we pre-train our StruMamba3D on ShapeNet dataset~\cite{chang2015shapenet}, which contains 52,472 unique 3D models across 55 common object categories.
For each shape, we sample 1024 points as input and partition them into 64 groups using FPS and KNN, with 60\% of the groups being randomly masked.
The pre-training process utilizes the AdamW~\cite{kingma2014adam} optimizer, incorporating cosine learning rate decay~\cite{loshchilov2016sgdr}. 
The initial learning rate is set to 1e-3, with a weight decay of 5e-2 and a dropout rate of 1e-1. 
The batch size is 128, and training is performed for 300 epochs.
The model pre-training is conducted on a single NVIDIA RTX A6000 GPU and takes approximately 18 hours to complete.

\subsection{Fine-tuning Details}

\textbf{Shape Classification on ScanObjectNN Dataset.}
ScanObjectNN is a challenging real-world dataset comprising approximately 15,000 objects across 15 distinct categories.
For the shape classification task, we first sample 2048 points from each object as input.
Next, we leverage our proposed StruMamba3D to extract discriminative group features, which are then aggregated through maximum and average pooling operations to obtain comprehensive global point cloud representations.
These global features are processed through a classification head consisting of three linear layers, with respective output dimensions of 256, 256, and $N_{cls}$ (the number of object categories).
Finally, a cross-entropy loss is applied to supervise the predictions.
\begin{equation}
    \label{crossEntropy}
    \mathcal{L}_{cls} = -\sum_{i=1}^{N_{cls}} y_{i} \log(p_{i}),
\end{equation}
where $y_i$ is the indicator function taking a value of 1 when the true class of the point cloud is $i$ and 0 otherwise, and $p_i$ denotes the predicted probability for class $i$.
For the fine-tuning process, the model is trained for 300 epochs with a batch size of 32 and a learning rate of 5e-4. 
We utilize the AdamW optimizer~\cite{kingma2014adam} with a weight decay of 5e-2 and apply cosine learning rate decay~\cite{loshchilov2016sgdr}, with an initial learning rate of 1e-6 and a warm-up period of 10 epochs.

\noindent\textbf{Shape Classification on ModelNet40 Dataset.}
ModelNet40 is a widely used benchmark dataset for 3D shape classification, containing 12,311 unique 3D models spanning 40 categories.
For the shape classification task, we begin by sampling 1,024 points from each object as input.
All other experimental settings remain consistent with those used for the ScanObjectNN dataset.
During testing, we report results both with and without the voting strategy.
The voting strategy involves conducting ten tests with random scaling and averaging the predictions to obtain the final result.

\noindent\textbf{Part Segmentation on ShapeNetPart Dataset.}
ShapeNetPart~\cite{chang2015shapenet} is a benchmark dataset specifically designed for 3D part segmentation. 
It comprises 16,881 unique 3D models spanning 16 object categories and 50 part categories, providing a diverse and comprehensive dataset for evaluating fine-grained 3D shape understanding.
For the part segmentation task, we begin by sampling 2,048 points from each shape as input.
Following PointBERT~\cite{yu2022pointbert}, we adopt a similar segmentation head architecture and extract features from the $4$-th, $8$-th, and $12$-th layers of our StruMamba3D.
These three levels of features are concatenated and then processed separately using average pooling and max pooling to obtain two distinct global features.
Besides, leveraging feature propagation from PointNet++~\cite{qi2017pointnet++}, we upsample the concatenated features to match the 2,048 input points, ensuring per-point feature representation.
The per-point features are then concatenated with the two global features and passed through a Multi-Layer Perceptron (MLP) for point-wise label prediction.
To optimize the model, we use cross-entropy loss to supervise the predictions, which is formulated as follows:
\begin{equation}
    \label{NLLLoss}
    \mathcal{L}_{seg} = -\frac{1}{N} \sum_{i=1}^{N} \sum_{j=1}^{N_{seg}} y^{i}_{j}  \log(p^{i}_{j}),
\end{equation}
where $N$ represents the number of points, $N_{seg}$ denotes the number of part categories, $y^{i}_{j}$ is the indicator function taking a value of 1 when the true class of the point $i$ is $j$ and 0 otherwise, and $p^{i}_{j}$ denotes the predicted probability for point $i$ being classified as $j$.
For fine-tuning, the model is trained for 300 epochs with a batch size of 32 and a learning rate of 2e-4. 
We utilize the AdamW optimizer~\cite{kingma2014adam} with a weight decay of 5e-2. 
Additionally, we apply cosine learning rate decay~\cite{loshchilov2016sgdr}, starting with an initial learning rate of 1e-6 and incorporating a warm-up period of 10 epochs to stabilize early training.

\noindent\textbf{Few-Shot Classification on ModelNet40 Dataset.}
In the $n$-way-$k$-shot setting, the support set consists of $n$ distinct classes, each containing $k$ samples.
The model is trained only using the sampled $n \times k$ samples. 
For evaluation, we randomly sample 20 novel instances from each of the $n$ classes to form the test set.
We conduct few-shot classification experiments on the ModelNet40 dataset across four different configurations: 5-way-10-shot, 5-way-20-shot, 10-way-10-shot, and 10-way-20-shot.
Each configuration is evaluated over 10 independent trials, and we report the mean accuracy along with the standard deviation.
The training hyperparameters remain consistent with those used in the shape classification experiments on the ModelNet40 dataset.

\section{More Experimental Results}\label{sec:exp}

\subsection{Detailed Results on Part Segmentation}

We present per-category part segmentation results on the ShapeNetPart~\cite{yi2016scalable} dataset.  
The results for PointGPT-S~\cite{chen2023pointgpt} and PointMamba~\cite{liang2024pointmamba} are reproduced using their official code.
As shown in~\cref{table:ShapeNetPart}, our method achieves the best performance in most categories.  
The segmentation task requires the model to effectively capture the local structure of point clouds.
Our approach enhances local structure modeling by introducing spatial states, which play a crucial role in improving segmentation performance. 
This spatial state modeling allows the network to better preserve spatial relationships and structural details, leading to superior results in segmentation tasks.
%

\begin{table*}[!t]
    \begin{center}
      \caption{\textbf{Detailed Results of Part Segmentation on ShapeNetPart Dataset.} We report the mean $\operatorname{IoU}$ for each category. * indicates results reproduced using the official code.}
      \small
      \label{table:ShapeNetPart}
      \begin{tabular}{l|cccccccccccccccccc}
        \toprule
        \multirow{1}*{Method} & $\text{mIoU}_c$ & $\text{mIoU}_i$ & airplane & bag & cap & car & chair & earphone & guitar \\
        \midrule
        MaskPoint~\cite{liu2022maskpoint} & \underline{84.6} & 86.0 & 84.2 & 85.6 & 88.1 & 80.3 & 91.2 & 79.5 & 91.9 \\
        PointBERT~\cite{yu2022pointbert} & 84.1 & 85.6 & 84.3 & 84.8 & 88.0 & 79.8 & 91.0 & \textbf{81.7} & 91.6 \\
        PointMAE~\cite{yu2022pointbert} & 84.2 & \underline{86.1} & 84.3 & 85.0 & 88.3 & \underline{80.5} & \underline{91.3} & 78.5 & \underline{92.1} \\
        PointGPT-S*~\cite{chen2023pointgpt} & 84.0 & 85.9 & \textbf{85.1} & \underline{86.1} & \underline{88.8} & 80.2 & 91.2 & 78.8 & 91.7 \\
        PointMamba*~\cite{liang2024pointmamba} & 83.8 & 85.9 & 84.5 & 84.7 & 87.8 & 80.2 & 91.2 & 78.7 & 91.9 \\
        \rowcolor{gray!20} 
        \textbf{Ours} & \textbf{85.0} & \textbf{86.7} & \underline{84.8} & \textbf{87.4} & \textbf{89.2} & \textbf{81.3} & \textbf{91.6} & \underline{80.8} & \textbf{92.2} \\
        \midrule
        \multirow{1}*{Method} & knife & lamp & laptop & motorbike & mug & pistol & rocket & skateboard & table \\
        \midrule
        MaskPoint~\cite{liu2022maskpoint} & 87.8 & \textbf{86.2} & 95.3 & \underline{76.9} & \underline{95.0} & \underline{85.3} & \textbf{64.4} & \underline{76.9} & 81.8 \\
        PointBERT~\cite{yu2022pointbert} & \underline{87.9} & 85.2 & 95.6 & 75.6 & 94.7 & 84.3 & 63.4 & 76.3 & 81.5 \\
        PointMAE~\cite{yu2022pointbert} & 87.4 & \underline{86.1} & \textbf{96.1} & 75.2 & 94.6 & 84.7 & \underline{63.5} & \textbf{77.1} & \textbf{82.4} \\
        PointGPT-S*~\cite{chen2023pointgpt} & 87.0 & 84.5 & \underline{95.9} & 74.2 & \underline{95.0} & 83.8 & 62.6 & 76.8 & 81.6 \\
        PointMamba*~\cite{liang2024pointmamba} & 87.3 & 85.1&95.8&74.0&94.6&84.7&61.6&75.9&\textbf{82.4}\\
        \rowcolor{gray!20} 
        \textbf{Ours} & \textbf{88.3} & {86.0} & {95.8} & \textbf{78.1} & \textbf{95.5} & \textbf{85.8} & \underline{63.5} & \underline{76.9} & \underline{82.2}\\
        \bottomrule
      \end{tabular}
    \end{center}
  \end{table*}

\subsection{Large-scale 3D Scene Task}
To evaluate the scalability of our approach to large-scale 3D scene understanding tasks, we conduct 3D object detection experiments on the ScanNetV2 dataset, which contains approximately 50K points per scene.
For fair comparison, we follow the evaluation protocol of Point-M2AE and adopt 3DETR-m as our baseline.
Our encoder is configured with the same number of layers as Point-M2AE, processing 2048 input tokens and employing 128 spatial states.
We pretrain our model on ScanNetV2 for 1080 epochs using a learning rate of 5e-4 and a batch size of 16. All other hyperparameters are kept consistent with those used in the ShapeNet experiments.
During fine-tuning, we strictly follow the 3DETR-m setup to ensure a fair comparison.
As shown in~\cref{table:ablation3}, the experimental results clearly demonstrate the effectiveness and generalization ability of our method in large-scale 3D scene settings.

\begin{table}[!t]
    \begin{center}
    \small
    \caption{\textbf{Quantitative comparison on the ScanNetV2 dataset in terms of $\text{mAP}{25}$ and $\text{mAP}{50}$.}}
    \label{table:ablation3}
    \begin{tabular}{l|cc}
        \hline
        Method & $\text{mAP}_{25}$ & $\text{mAP}_{50}$ \\
        \hline
        3DETR & 62.1 & 37.9 \\
        3DETR-m & 65.0 & 47.0 \\
        Point-M2AE & 66.3 & 48.3 \\
        Ours & \textbf{67.8} & \textbf{50.6} \\
        \hline
    \end{tabular}
    \end{center}
\end{table}

\subsection{Ablation of Structural Modeling}
Unlike the standard Mamba, we use hidden states to model the structural information of point clouds.
During the state update and propagation process, we incorporate the relative positions between input points and states into the generation of SSM parameters. 
This spatial interaction enables each state to selectively focus on and update the features of points within its designated region.
As a result, StruMamba3D eliminates the need for serialization strategies to reorder the point cloud.
To further evaluate the impact of our proposed structural modeling and point serialization strategies, we present the experiments in~\cref{table:ablation_ss}.
For the baseline using the standard Mamba block, adding serialization improves model performance.
However, due to the distortion of spatial adjacency, the performance still lags behind our method (89.87 vs. 92.75 on ScanObjectNN, 93.68 vs. 95.06 on ModelNet40, and 84.07 vs. 84.96 on ShapeNetPart).
StruMamba3D achieves promising performance without the need for serialization, and adding serialization provides only a modest performance gain. 
Given the computational overhead introduced by serialization, we choose to remove it from StruMamba3D.
\begin{table}[!t]
    \begin{center}
        \small
        \setlength\tabcolsep{5pt}
        \caption{\textbf{Ablation of Structural Modeling.}}
        \label{table:ablation_ss}
        \begin{tabular}{l|c|ccc}
            \toprule
            \multirow{2}*{Backbone} & \multirow{2}*{Serialization} &\multicolumn{2}{c}{Overall Accuracy} & mIoU$_c$  \\
            \cline{3-5}                 
            ~ & ~ & ScanNN & MN40 & SNPart \\
            \hline
            Mamba & \ding{55} & 88.24 & 92.50 & 82.08  \\
            Mamba & Z-order & 89.28 & 93.44 & 83.89 \\
            Mamba & Hilbert & 89.87 & 93.68 & 84.07 \\
            StruMamba3D & \ding{55} & \underline{92.75} & \textbf{95.06} & \underline{84.96} \\
            StruMamba3D & Z-order & {92.47} & \underline{94.98} & {84.86} \\
            StruMamba3D & Hilbert & \textbf{92.81} & {94.89} & \textbf{85.12} \\
            \bottomrule
        \end{tabular}
    \end{center}
\end{table}

\subsection{Ablation of the Number of Spatial States}
In this work, we introduce spatial states into the SSM to capture the structural information of point clouds.
To investigate the impact of the number of spatial states on model performance, we conduct ablation studies as shown in~\cref{table:ablation1}.
The results show that reducing the number of spatial states leads to a performance drop, highlighting the importance of spatial states in capturing point cloud structures.
When the number of spatial states exceeds 16, the performance no longer improves significantly with further increases.  
We believe that 16 spatial states are sufficient to represent the structural information of objects.  
Therefore, we set $M$=16 as the number of spatial states to balance model performance and computational cost.
\begin{table}[!t]
    \begin{center}
        \small
        \caption{\textbf{Ablation of the number $M$ of spatial states.} }
        \label{table:ablation1}
        \begin{tabular}{c|ccc}
            \toprule
            \multirow{2}*{ $M$} & \multicolumn{2}{c}{Overall Accuracy}  & $\text{mIoU}_c$    \\
            \cmidrule{2-4}                 
            ~ & ScanNN & MN40 & SNPart\\
            \midrule
            4 & 91.74 & 93.35 & 84.34 \\
            8 & 92.26 & \underline{94.89} & 84.71 \\
            16 & \textbf{92.75} & \textbf{95.06} & \underline{84.96} \\
            24 & \underline{92.57} & 94.81 & \textbf{85.07}\\
            \bottomrule
        \end{tabular}
    \end{center}
\end{table}

\subsection{Ablation on the Lightweight Convolution}
In the structural SSM block, we propose a lightweight convolution for spatial states and input points.
This design offers two key benefits: (1) Enhance feature interactions among spatial states. (2) Replace the causal conv1d in the original SSM block.
In lightweight convolution, the number of neighboring points determines the receptive field of the module.
To select an appropriate number of neighbors, we conduct an ablation study on the number of neighbors for spatial states and input points. 
As shown in~\cref{table:ablation2}, the model achieves optimal performance when $k$=4 for spatial states and $k$=8 for input points.
Additionally, we observe that the model is more sensitive to the number of neighbors of spatial states than to that of input points.
This observation suggests that spatial states play a more crucial role in capturing structural information.
\begin{table}[!ht]
    \begin{center}
        \small
        \caption{\textbf{Ablation on the point convolution.} ``$k$'' denotes the number of k-nearest neighbors in the lightweight convolution.}
        \label{table:ablation2}
        \begin{tabular}{cc|ccc}
            \toprule
            \multirow{2}*{$k$ for $F_h$} & \multirow{2}*{$k$ for $F_x$} & \multicolumn{2}{c}{Overall Accuracy}   & $\text{mIoU}_c$   \\
            \cmidrule{3-5}                 
            ~ & ~ & ScanNN & MN40 & SNPart\\
            \midrule
            4 & 8 & \textbf{92.75} & \textbf{95.06} & \textbf{84.96} \\
            2 & 8 & 91.91 & 94.29 & 84.54 \\
            8 & 8 & 92.71 & 94.98  & 84.92 \\
            4 & 4 & 92.30 & 94.65 & 84.71 \\
            4 & 12 & 92.37 & 94.69 & 84.86 \\
            \bottomrule
        \end{tabular}
    \end{center}
\end{table}

\subsection{Linear Complexity of StruMamba3D}
\begin{figure}[!t]
    \begin{center}
        \includegraphics[width=0.4\textwidth]{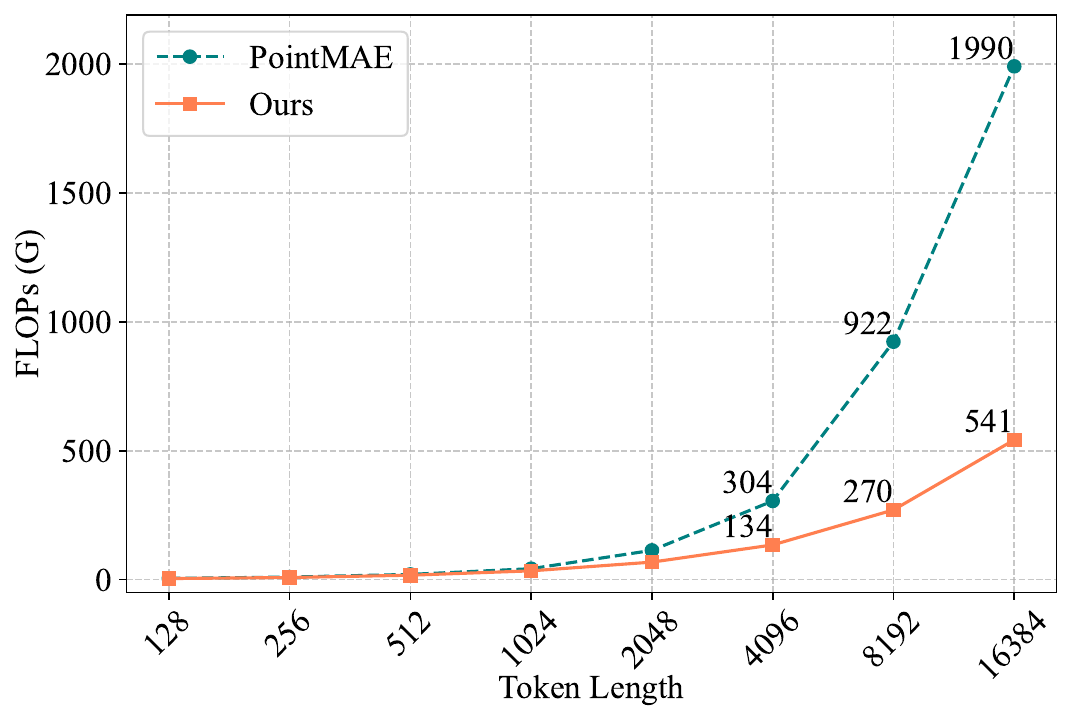}
        \vspace{-1em}
        \caption{\textbf{FLOPs comparison with PointMAE.} }
        \label{fig:flops_tendency}
    \end{center}
\end{figure}

\begin{figure*}[!t]
    \centering
    \includegraphics[width=0.8\textwidth]{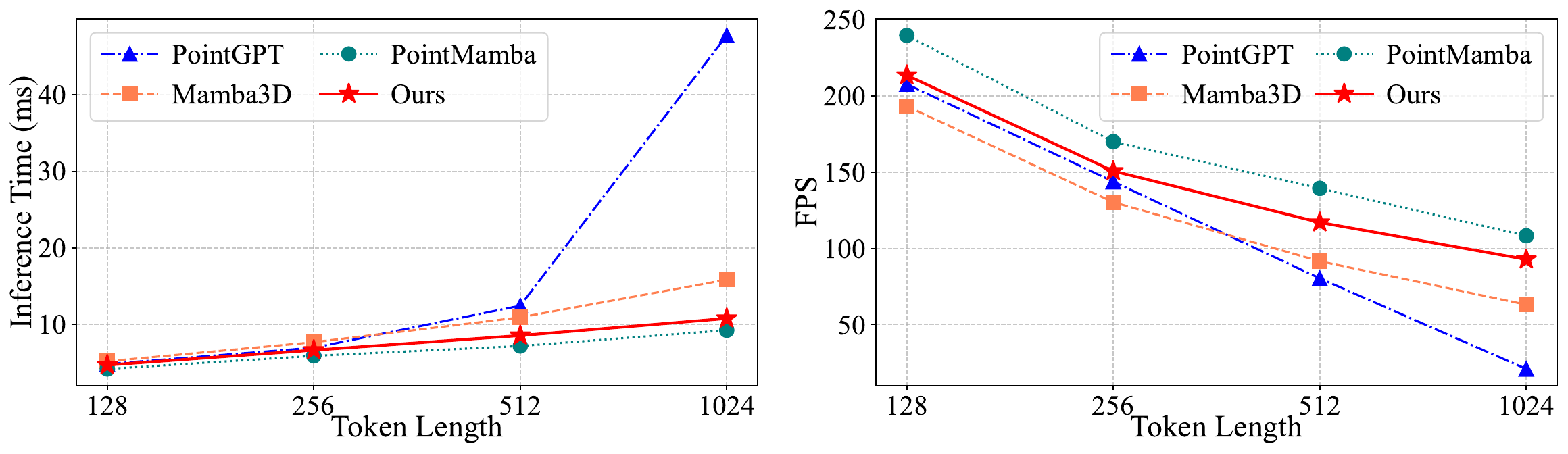}
    \label{fig:speed_comparison}
    \caption{\textbf{Inference time and FPS comparison.}}
\end{figure*}

Leveraging hardware-aware scan algorithm of Mamba~\cite{gu2023mamba}, StruMamba3D exhibits linear computational complexity.  
As illustrated in~\cref{fig:flops_tendency}, we present a FLOPs comparison with the transformer-based method PointMAE, where StruMamba3D demonstrates significantly lower computational cost when processing long sequences.  
%
%
We also report the inference time and FPS under different input lengths.
As shown in~\cref{fig:speed_comparison}, our method is faster than the Transformer-based method PointGPT, especially with longer input lengths.
Besides, our method also achieves competitive inference speed compared to other Mamba-based methods.

\subsection{Performance without Pretraining}
We provide the results of training our model from scratch (without pretraining).
As shown in following table, our method still performs well, outperforming existing supervised learning only methods.
\begin{table}[!t]
    \begin{center}
        \small
        \setlength\tabcolsep{2pt}
        \label{table:ablation1}
        \begin{tabular}{l|ccc}
            \hline
            \multirow{2}*{Method} & ScanObjectNN & ModelNet40  & {ShapeNetPart}      \\
            ~ & mOA & mOA & $\text{mIoU}_{c}$ \\
            \hline
            w/o Pretraining & 91.33 & 93.68 & 83.96 \\
            MPM Pretraining & 92.09 & 94.45 & 84.49 \\
            Our Pretraining & \textbf{92.75} & \textbf{95.06}  & \textbf{84.96}\\
            \hline
        \end{tabular}
    \end{center}
  \end{table}

\section{Qualitative Analysis}\label{sec:visual}

\subsection{Visualization of Spatial State Correlation}
\begin{figure*}[!t]
    \vspace{-0.5em}
    \begin{center}
       \includegraphics[width=0.9\textwidth]{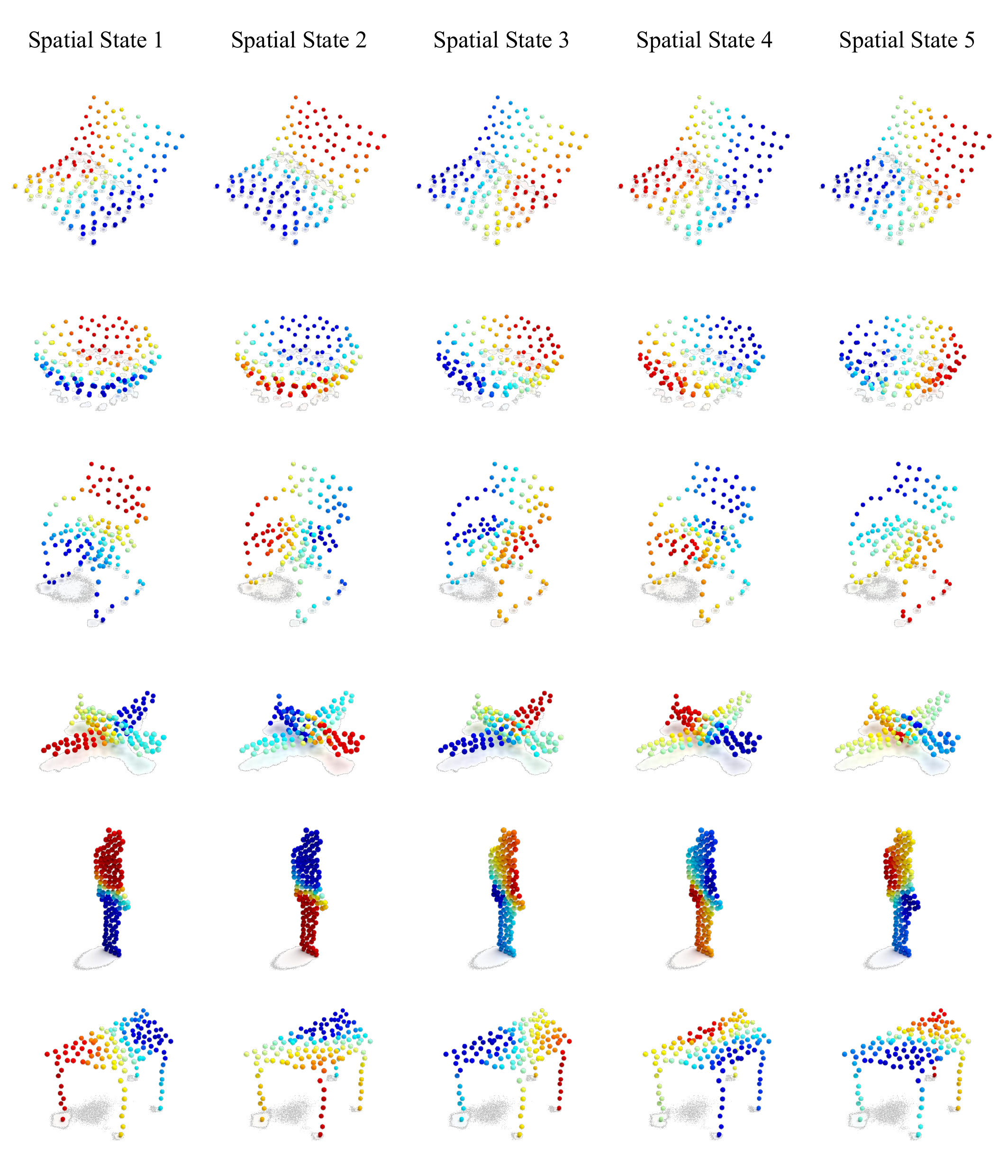}
       \caption{\textbf{Visualization of correlation between the output point and spatial state features produced by the structural SSM.} For each point cloud, we select five spatial states that are widely spaced for visualization.
 }
       \label{fig:spatial_state_correlation}
    \end{center}
\end{figure*}

We introduce spatial states into the SSM to capture the local structure of point clouds, thereby preserving the spatial dependencies among points.
To validate the effectiveness of spatial states, we visualize the correlations between the output point features and spatial state features produced by the structural SSM.
Since neighboring spatial states may focus on the same region, we select five spatial states that are widely spaced for visualization.
As shown in~\cref{fig:spatial_state_correlation}, different spatial states focus on different parts of the point cloud.
The results demonstrate that our structural SSM can effectively capture the local structure of point clouds by spatial states.

\subsection{Visualization of Masked Point Modeling}
\begin{figure*}[!t]
    \begin{center}
       \includegraphics[width=0.99\textwidth]{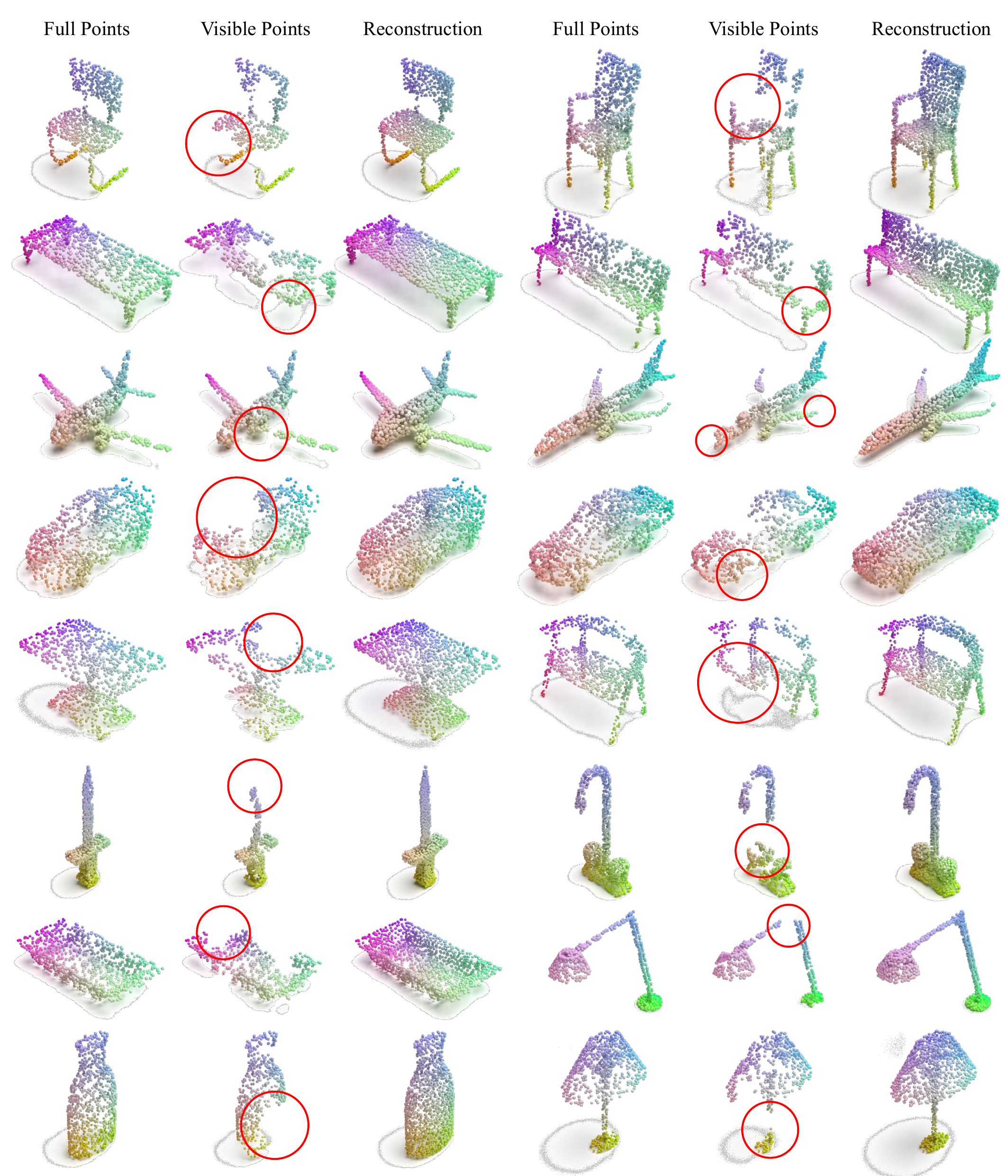}
       \caption{\textbf{Visualization of reconstructed masked regions on ShapeNet.} Full points represent the raw point cloud, while visible points correspond to the input points with 60\% masked. The reconstruction results consist of the predicted points combined with visible points.
 }
       \label{fig:masked_modeling}
    \end{center}
\end{figure*}

In the pretraining phase, we adopt masked point modeling as our primary self-supervised learning objective.
Given a point cloud, we randomly mask 60\% of the point groups and use StruMamba3D to extract features, followed by a shallower decoder that predicts the coordinates of masked points.
To further analyze the ability of our model to perceive structural information, we visualize the reconstruction results on the ShapeNet dataset.
As shown in~\cref{fig:masked_modeling}, our model can effectively reconstruct the original structure even when 60\% of the points are masked.
Moreover, our model performs remarkably in reconstructing complex geometric patterns, such as the delicate shapes of chair backs and table legs.
These visualization results validate that our approach can effectively capture and encode both local and global structural information from point clouds, which is crucial for downstream tasks.

\subsection{Visualization of Part Segmentation}
\begin{figure*}[!t]
    \begin{center}
       \includegraphics[width=0.99\textwidth]{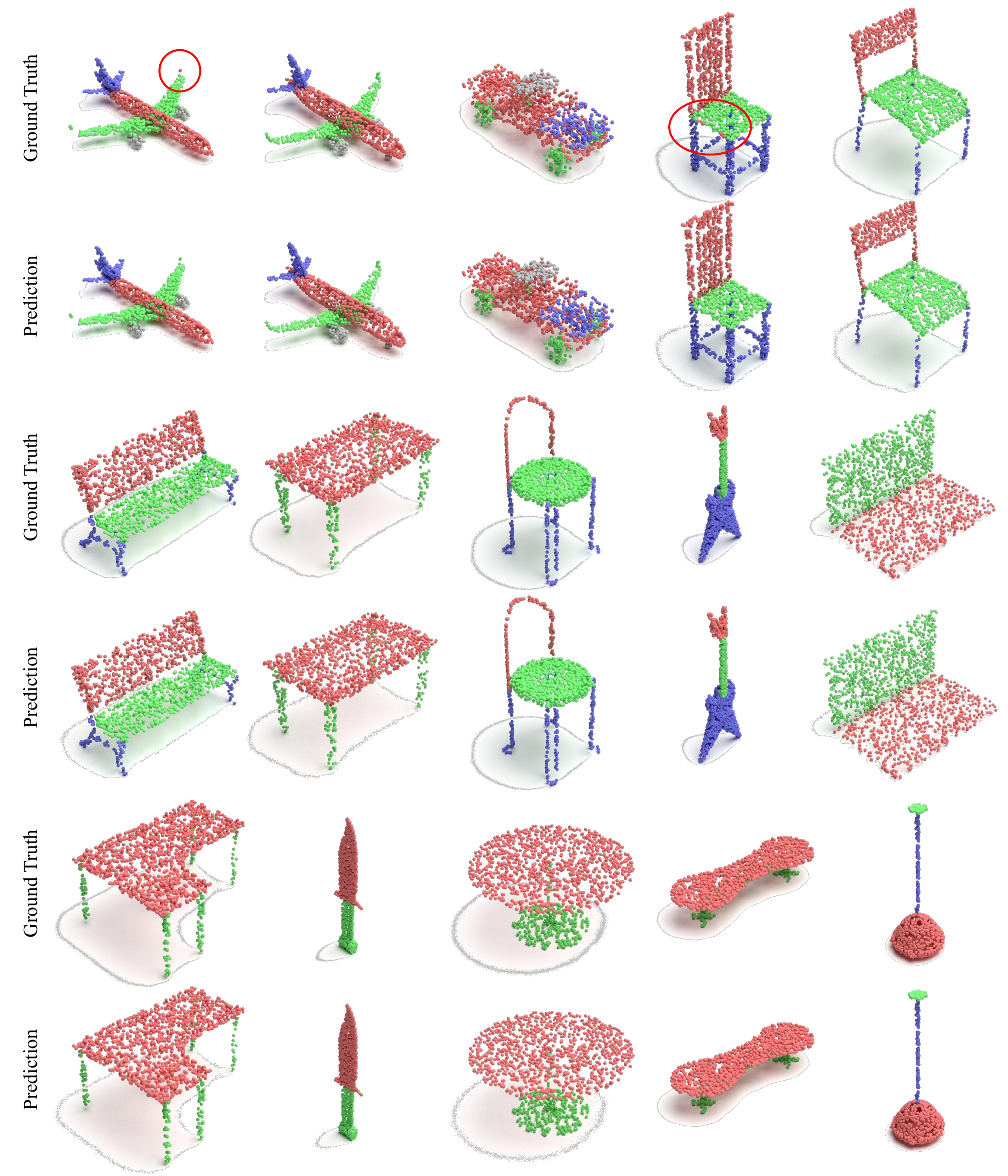}
       \caption{\textbf{Visualization of part segmentation of our Structural Mamba on ShapeNetPart.} Different colors represent different parts. The red circles highlight points with obvious annotation errors.
 }
       \label{fig:part_seg}
    \end{center}
\end{figure*}

In this subsection, we present the qualitative results of part segmentation on the ShapeNetPart validation set, comparing ground truth and predictions.
As shown in~\cref{fig:part_seg}, our method demonstrates highly competitive performance in part segmentation.
Notably, the ShapeNetPart dataset contains certain annotation errors, as highlighted by the red circles.
Nevertheless, our method not only achieves a high mIoU but also correctly segments points with erroneous annotations.
Moreover, for complex structures such as wheels and chair legs, our approach delivers accurate and reliable segmentation results.

\section{Discussions}\label{sec:discussion}

In this section, we discuss the limitations of our work and potential directions for future research.
Our primary goal is to fully exploit the potential of Mamba for point cloud representation learning.
To address two key issues of Mamba: disrupting the spatial adjacency of point clouds and struggling to maintain long-sequence memory in downstream tasks, we propose the structural SSM block and the sequence length-adaptive strategy, respectively.
While our approach effectively models the structural information of point clouds and achieves significant performance improvements across four downstream tasks, certain challenges remain.
Specifically, in part segmentation tasks, we observe some results with unclear boundaries.
We attribute this limitation to the nature of single-scale models, which focus on fixed-scale features and struggle to capture geometric details across multiple scales.
For example, larger structures such as chair backs and finer details like table legs require different scale features for precise differentiation.
To address this, a promising future direction is the development of a hierarchical Mamba that can perform multi-scale point cloud feature extraction through spatial state modeling. 
This could enhance the ability of the model to adaptively capture both global structures and fine-grained details, leading to more accurate segmentation results.

\clearpage

\end{document}